# 2D Face Recognition System Based on Selected Gabor Filters and Linear Discriminant Analysis LDA

Samir F. Hafez[1], Mazen M. Selim[2] and Hala H. Zayed[3]

[1] PhD Student, Benha University, Faculty of Engineering
Benha University, Benha, Egypt
HafezSF@Gmail,com

[2] Department of computer Science, Faculty of computers and informatics,
Benha University, Benha, Egypt
Selimm@bu.edu.eg

[3] Department of computer Science, Faculty of computers and informatics,
Benha University, Benha, Egypt
Hala.zayed@fbci.bu.edu.eg

**Abstract**

In this paper, we present a new approach for face recognition system. The method is based on 2D face image features using subset of non-correlated and Orthogonal Gabor Filters instead of using the whole Gabor Filter Bank, then compressing the output feature vector using Linear Discriminant Analysis (LDA). The face image has been enhanced using multi stage image processing technique to normalize it and compensate for illumination variation. Experimental results show that the proposed system is effective for both dimension reduction and good recognition performance when compared to the complete Gabor filter bank. The system has been tested using CASIA, ORL and Cropped YaleB 2D face images Databases and achieved average recognition rate of 98.9 %.

## 1. Introduction

Automatic human face recognition has received substantial attention from researchers in biometrics, pattern recognition, and computer vision communities over the past few decades for solving its outstanding challenges which exist in the uncontrolled environment [1], [2]. It does not require the cooperation of individual with the system so that it is intuitive and nonintrusive method for recognizing people. Current 2D face recognition systems still encounter difficulties in handling facial variations due to head poses, lighting conditions and facial expressions [3], which introduce large amount of intra-class variations. Range images captured by a 3D sensor explicitly contain facial surface shape information. The 3D shape information does not change much due to pose and lighting variations, which can change the corresponding intensity image significantly. Range image based 3D face recognition has been demonstrated to be effective in enhancing the face recognition accuracy [4], [5].

In this paper, we focus on face recognition using 2D face images. A number of approaches has been developed for face recognition using 2D Images. Turk and Pentland [1] proposed eigenfaces based Face recognition, where images are projected onto a feature space ("face space") that best encodes the variation among known face images. The face space is defined by the "eigenfaces", which are the eigenvectors of the set of faces. Omai, et al. [2] proposed system based on distance between the Discrete Cosine Transform (DCT) of the face under evaluation and all the DCTs of the faces database, the face with the shortest distances probably belong to the same person. P. Yang, et al. [3] has applied AdaBoost Gabor feature which are low dimensional and discriminant. Bouzalmat, et al. [4] have proposed system based on the Fourier transform of Gabor filters and the method of regularized linear discriminate analysis applied to facial features previously localized. The process of facial face recognition is based on two phases: location and recognition. The first phase determines the characteristic using the local properties of the face by the variation of gray level along the axis of the characteristic and the geometric model, and the second phase generates the feature vector by the convolution of the Fourier transform of 40 Gabor filters and face, followed by application of the method of regularized linear discriminate analysis on the vectors of characteristics.

Linlin SHEN [6] used Support Vector Machine (SVM) face identification method using optimized Gabor features, 200 Gabor features are first selected by a boosting algorithm which are then combined with SVM to build a two-class based face recognition system. Liu and Wechsler [5] Apply independent Gabor features (IGFs) method and its application to face recognition. The novelty of the IGF method comes from 1) the derivation of independent Gabor features in the feature extraction stage and 2) the development of an IGF features-based probabilistic reasoning model (PRM) classification method in the pattern recognition stage. In particular, the IGF method first derives a Gabor feature vector from a set of down-sampled Gabor wavelet representations of face images, then reduces the dimensionality of the vector by means of principal component analysis, and finally defines the independent Gabor features based on the independent component analysis (ICA). The independence property of these Gabor features facilitates the application of the PRM method for classification. Sang-Il Choi, Chong-Ho Choi and Nojun Kwakb [7] proposed a novel 2D image-based approach that can simultaneously handle illumination and pose variations to enhance face recognition rate. It is much simpler, requires much less computational effort than the methods based on 3D models, and provides a comparable or better recognition rate. Kyungnam Kim [8] proposed a face recognition system based on Principle Component Analysis (PCA) , which computes the basis of a space which is represented by its training vectors. These basis vectors, actually eigenvectors, computed by PCA in the direction of the largest variance of the training vectors. Each eigenface can be viewed a feature. When a particular face is projected onto the face space, its vector into the face space describe the importance of each of those features in the face. The face is expressed in the face space by its eigenface coefficients (or weights).

This paper presents a novel technique for 2D face recognition based on feature extraction of 2D face image using selected Gabor filters which best describe the face image by selecting the most discriminative features that represent the face image. Gabor filters are promoted for their useful properties, such as invariance to illumination, rotation, scale and translations, in feature extraction [9]. The use of the part of Gabor filter bank can effectively decrease the computation and reduce the dimension, even improve the recognition capability in some situations. For further dimensionality reduction and good recognition performance we used PCA for feature compression and selection.
The remainder of the paper is organized as follows: Section 2 describes the preprocessing procedure to enhance and normalize the face image. Section 3 presents the Gabor feature extraction and selecting the best Gabor filter bank that represent the face image. Feature compression based on PCA discussed in Section 4. In Section 5, experiments are performed using the Chinese Academy of Sciences - Institute of Automation (CASIA) 2D face Dataset [10] and a. AT&T Database of Faces, (formerly 'The ORL Database of Faces) [11] and Cropped YaleB face database [12]. Finally, conclusion is given in Section 6.

## 2. Preprocessing

Preprocessing procedure is very important step for 2D face recognition system.
 The goal of processing is to obtain normalized face images, which have normalized intensity, uniform size and shape. It also should eliminate the effect of illumination and lighting. Our proposed preprocessing technique consists of three steps, Automatic landmark detection, Face image normalization and finally Illumination compensation.

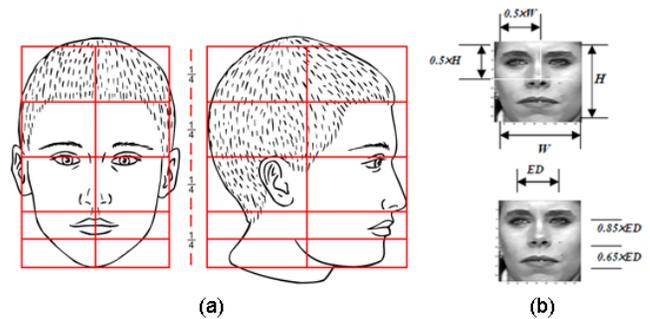

Fig. 1 Human Face Geometry

2.1 Automatic landmark detection

Human face has a very distinctive structure, where human face can be divided into six equal squares [13] as shown in Figure (1-a) . The structure of the human face has been used to divide the face regions into separate sets of regions each one contains one or more landmarks to search for it. We need to set an Anchor for the face image to start dividing it into separate ROIs, this anchor can be the center of the eyes where nose and mouth locations are related to the eye distance ED (Distance between Center of the Eyes) as shown on Figure (1-b). Detection of the center of the eyes can be accomplished by using template matching algorithm based 2D Normalized Cross Correlation 2DNCC [14]. 2DNCC compensate the brightness variation between image and template due to lighting and exposer conditions by normalizing the image at every step. This is typically done by subtracting the mean and dividing by the standard deviation. Cross-Correlation of a template, with a sub-image is:

$$c(t) = \frac{1}{n-1} \sum_{x,y} \frac{(f(x,y)-\bar{f})(t(x,y)-\bar{t})}{\sigma_f \sigma_t} \quad (1)$$

Where n is the number of pixels in t(x, y) and f(x, y), $\bar{f}$ is the average of *f*. and $\sigma_f$ is standard deviation of *f*. The cross correlation output $c(t)$ achieves its maximum value of 1 if and only if the template matches the sub-image.

2D Normalized Cross Correlation has been used to detect the location of the center of eyes in the upper half of the face. Then, the detected eyes center coordinates are used to compute the Region of interests. We have created templates manually for the eye to be used by the 2D Normalized cross correlation which gave very good detection results.

2.2 Face Image Normalization

The next step after eye center coordinates detection is to extract the facial region from an image. Then normalize the image in terms of geometry and size (rotation, cropping and scaling). The same process will be applied in case of color images. We can define the mouth region to be the horizontal strip whose top is at 0.85×ED from the eyes horizontal position and has a height equal to 0.65×ED [15] as shown in Figure (1-b)

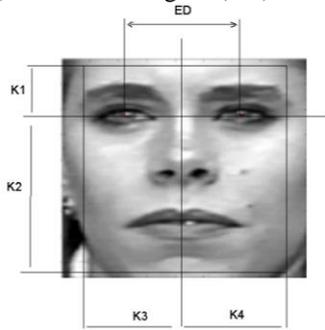

Fig. 2 Normalized face image

We extract the facial region from the given image based on the eye coordinates and eye distance ED as shown in Figure 2. The image is then normalized to a fixed size given by the parameter 128x128 where:
K1=ED
K2=2*ED
K3=1.3*ED
K4=1.3*ED

2.3 Illumination Compensation

The variability on the 2D face images brought by illumination changes is one of the biggest obstacles for reliable and robust face verification. Research has shown that the variability caused by illumination changes can easily exceeds the variability caused by identity changes [16]. Illumination normalization, therefore, is a very important topic to study. Illumination preprocessing on 2D images can be divided into two groups: Histogram Normalization and photometric Normalization [17].

Photometric Normalization based on the idea that image *I(x,y)* is the product of two components, illumination *L(x,y)* and reflectance *R(x,y)* , introduce in 1971 by Land et al. [17].

$$I(x,y) = L(x,y) * R(x,y) \quad (2)$$

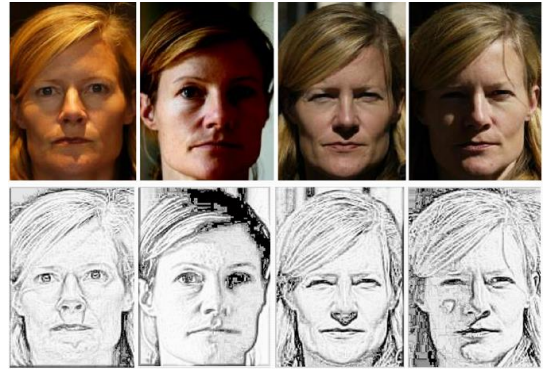

Fig. 3 ASR Result on different Illuminations

Illumination contains geometric properties of the scene (i.e., the surface normals and the light source position) and reflectance contains information about the object. Based on the assumption that the illumination varies slowly across different locations of the image and the local reflectance may change rapidly across different location, the processed illumination should be drastically reduced due to the high-pass filtering, while the reflectance after this filtering should still be very close to the original reflectance. The reflectance can be also finding by dividing the image by the low pass version if the original image, which is representing illumination components.

Adaptive Single Scale Retinex (ASR) has been used to deal with  'graying out' , 'noise enlargement' and 'halo'

effects existing in classical Retinex image enhancement algorithms. ASR was presented by Park in [18]. The proposed method estimates illumination by iteratively convolving the input image with a 3× 3 smoothing mask weighted by a coefficient via combining two measures of the illumination discontinuity at each pixel.

## 3. Feature Extraction using Gabor

The Gabor filters, whose kernels are similar to the 2D receptive field profiles of the mammalian cortical simple cells, have been considered as a very useful tool in computer vision and image analysis due to its optimal localization properties in both spatial analysis and frequency domain [19].

### 3.1 Gabor Filters

Gabor filters has been found to be particularly appropriate for texture representation and discrimination. From information-theoretic view- point, Okajima [19]. derived Gabor functions as solutions for a certain mutual-information maximization problem. It shows that the Gabor receptive field can extract the maximum information from local image regions. Researchers have also shown that Gabor features, when appropriately designed, are invariant against translation, rotation, and scale [20].

Gabor filter is a linear filter used for edge detection In the spatial domain [21], a 2D Gabor filter is a Gaussian kernel function modulated by a sinusoidal plane wave. The filter has a real and an imaginary component representing orthogonal directions. The two components may be formed into a complex number or used individually.

Real

$$g(x,y;\lambda,\theta,\psi,\sigma,\gamma) = \exp\left(-\frac{x'^2+\gamma^2 y'^2}{2\sigma^2}\right)\cos\left(2\pi\frac{x'}{\lambda}+\psi\right) \quad (3)$$

Imaginary

$$g(x,y;\lambda,\theta,\psi,\sigma,\gamma) = \exp\left(-\frac{x'^2+\gamma^2 y'^2}{2\sigma^2}\right)\sin\left(2\pi\frac{x'}{\lambda}+\psi\right) \quad (4)$$

Where

$$x' = x\cos\theta + x\sin\theta$$

And

$$y' = -x\sin\theta + y\cos\theta$$

and $\lambda$ represents the wavelength of the sinusoidal factor, $\theta$ represents the orientation of the normal to the parallel stripes of a Gabor function, $\psi$ is the phase offset, $\sigma$ is the sigma of the Gaussian envelope and $\gamma$ is the spatial aspect ratio, and specifies the ellipticity of the support of the Gabor function.

Figure 4 shows the real part of the Gabor filters with 5 frequencies and 8 orientations for $\omega max=\pi/2$, the row corresponds to different frequency $\omega m$, the column corresponds to different orientation $\theta n$.

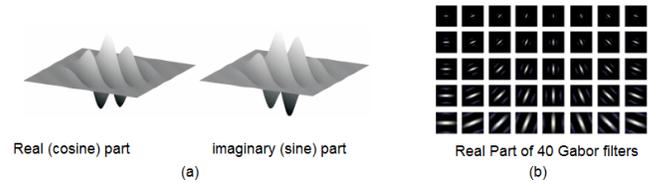

Real (cosine) part    Imaginary (sine) part     Real Part of 40 Gabor filters
(a)                                        (b)

Fig. 4- (a) Gabor filters Real and Imaginary parts-(b) 40 Gabor filters of 5 frequencies and 8 orientations

When exploited for feature extraction, a filter bank with several filters is usually created and used to extract multi-orientation and multi-scale features from the given face image. This filter bank commonly consist of Gabor filters of 5 different scales and 8 orientations [9].

Gabor features representation of the face image is the result of image I (x, y) convolution with the bank of Gabor filters $g_{u,v}(x, y)$. The convolution result is complex value which can be decomposed to real and imaginary part:

$$G_{u,v}(x, y) = I(x, y) * g_{u,v}(x, y) \quad (5)$$

$$E_{u,v}(x, y) = Re[G_{u,v}(x, y)] \quad (6)$$

$$O_{u,v}(x, y) = Im[G_{u,v}(x, y)] \quad (7)$$

Based on the decomposed filtering result both the phase $\varphi u,v(x, y)$ as well as the magnitude $A u,v(x, y)$ filter responses can be computed as:

$$A_{u,v}(x,y) = \sqrt{E^2_{u,v}(x,y) + O^2_{u,v}(x,y)} \quad (8)$$

$$\emptyset_{u,v}(x,y) = \arctan\left(\frac{O_{u,v}(x,y)}{E_{u,v}(x,y)}\right) \quad (9)$$

Since the computed phase responses vary significantly even for spatial locations only a few pixels apart, Gabor phase features are considered unstable and are usually discarded. The magnitude responses, on the other hand, varying slowly with the spatial position, and are thus the preferred choice when deriving Gabor filter based features.

## 3.2 Gabor Filter representation using Principle Component Analysis PCA

The Gabor face representation from a given face image I (x, y) is derived by using a bank of Gabor filters of 8 orientations and 5 scales as commonly used. , This procedure results in an inflation of the original pixel space to 40 times. For an image of size 64x64 pixels, 163840 features are used to represent it which is very large size. The issues storage space and computations are as important as the accuracy of verification and/or identification, so we implemented a technique to minimize the number of filters used by removing the redundancy and correlation between filters using Principle Component Analysis PCA.

Principal Component Analysis (PCA) is a well-known feature extraction method widely used in the areas of pattern recognition, computer vision and signal processing. PCA has been widely investigated and has become one of the most successful approaches in face recognition [23]. Principle Component Analysis (PCA) has been used to minimize the number of used filters by removing redundancy and correlation between them.

In the PCA-based face recognition methods, 2D face image matrices must be previously transformed into 1D image vectors column by column or row by row which often leads to a high-dimensional vector space. The vector space of 2D face image of size n*m is converted to 1D vector of length t=n*m;
Let us consider a set of N sample images $\{x_1, x_2, …, x_N\}$ represented by t-dimensional 1D face vector .
The PCA [23] can be used to find a linear transformation mapping the original t-dimensional feature space into an f-dimensional feature subspace, where normally f<<t. The new feature vector are defined by

$$y_i = W_{pca}^T x_{i(i=1,2,…,N)} \qquad (10)$$

Where, W$pca$ is the linear transformations matrix, i is the number of sample images.
The columns of W$pca$ are the *f* eigen vectors associated with the *f* largest eigen values of the covariance matrix$S_T$, which is defined as

$$S_T = \sum_{i=1}^{N}(x_i - \mu)(x_i - \mu)^T \qquad (11)$$

Where, µ is the mean image of all samples, the eigenvectors corresponding to nonzero eigenvalues of the covariance matrix$S_T$, produce an orthonormal basis for the subspace within which most image data can be represented with a small amount of error. The eigenvectors are sorted from high to low according to their corresponding eigenvalues. The eigenvector associated with the largest eigenvalue is one that reflects the greatest variance in the image. That is, the smallest eigenvalue is associated with the eigenvector that finds the least variance. They decrease in exponential fashion, meaning that the roughly 90% of the total variance is contained in the first 5% to 10% of the dimensions. Thus, number of Gabor filters used can be minimized by removing the redundancy and correlation between the applied filters and chooses number of filters that hold the total variance of the bank.

## 3.3 Gabor Filter Compression

Gabor filters are not orthogonal and hence correlated to each other's, PCA is used to produce linear orthogonal combination of the original Gabor filters with minimum correlation according to the properties of the PCA [2]. These generated filters are the eigenvectors of covariance data matrix.
The original Gabor filters bank has been used to form complex data matrix where each column represent one filter rearranged in 1D vector form of size abx1, where a,b are the size of the filter.
Consider G=[g1,g2,g3 ……g40] is complex data matrix contains 40 filters in vectors form.

$$G_T = \sum_{i=1}^{40}(g_i - \mu)(g_i - \mu)^T \qquad (12)$$

Where $G_T$ the 40x40 covariance matrix of the Gabor bank is, µ is the mean value. The eigenvectors corresponding to nonzero eigenvalues of the covariance matrix produce an orthonormal basis for the subspace within which most filter data can be represented with a small amount of error. Figure-(5.a) shows eigenvalues of the generated non-correlated filters.

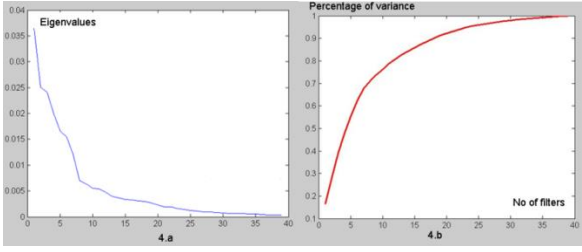

Fig. (5.a) Eigen Values      (5.b) percentage of variance

These eigenvectors are the principle components of the generated uncorrelated data set and can be used to generate a new filter bank representation by reshaping each eigenvector (abx1) size to a 2D filter of axb size. The first principle component shows the most dominant features of the dataset and succeeding components in turn shows the next possible dominant. The generated filters (principle components) are arranged in descending order according to their eigenvalues which is illustrated in Figure-(5-a). If we use all the generated filters then we are using the same Gabor bank in but orthogonal with minimum correlation form. Hence selecting subset of the generated filters according of percentage of variance as shown in Figure-(5.b) we select part of the filters which represent part of the total variance. Selecting the first 20 principle component (50% reduction in filter bank) will contribute 92.33% of the total variance and selecting 30 principle components (25 % reduction in filter bank) will contribute 98.01 % of the total variance. This process will decrease the computation and reduce the dimension of the generated feature vectors representation of the face image.

## 4. Feature Vector Compression using LDA

Although selecting subset of the Gabor filters bank will reduce the dimension of the data representation but we still have large data size, for a 20 orthogonal filters with an image of size 64x64, we have 81920 features. Using of Linear Discriminant Analysis (LDA) [23], for further compression of the features dataset. LDA seeks a projection that best separates the data in a least-squares sense. The main goal of LDA is to perform dimensionality reduction while preserving as much of the class discriminatory information as possible. It seeks to find directions along which the classes are best separated. While it takes into consideration the scatter within-classes but also the scatter between-classes.

4.1 Linear Discriminant Analysis LDA

Linear Discriminant Analysis (LDA) searches for those vectors in the underlying space that best discriminate among classes (rather than those that best describe the data). More formally, given a number of independent features relative to which the data is described, LDA creates a linear combination of these which yields the largest mean differences between the desired classes. Mathematically speaking, for all the samples of all classes, we define two measures:

1) One is called within-class scatter matrix, as given by

$$S_w = \sum_{j=1}^{c} \sum_{i=1}^{N_j} (x_i^j - \mu_j)(x_i^j - \mu_j)^T \quad (13)$$

Where $x_i^j$ is the ith sample of class j, $\mu_j$ is the mean of class j, c is the number of classes, and Nj the number of samples in class j; and

2) The other is called between-class scatter matrix

$$S_b = \sum_{j=1}^{c} (\mu_j - \mu)(\mu_j - \mu)^T \quad (14)$$

Where µ represents the mean of all classes.

The goal is to maximize the between-class measure while minimizing the within-class measure. One way to do this is to maximize the ratio $\frac{\det|S_b|}{\det(S_w)}$. The advantage of using this ratio is that it has been proven [23] that, if Sw is a nonsingular matrix then this ratio is maximized when the column vectors of the projection matrix, W, are the eigenvectors of $S_w^{-1} S_b$. It should be that: 1) there are at most c-1 nonzero generalized eigenvectors and, so, an upper bound on f is c-1, and 2) we require at least t+c samples to guarantee that Sw does not become singular (which is almost impossible in any realistic application).

## 5. Experimental Results

Three popular face image databases has been used for assessing the proposed system, CASIA 2D Face Database [10], ORL Database [11] and cropped yaleB database [12].

### CASIA 2D Face Database
Automation free download data set for 123 persons with two sets, intensity and range. Each person has 38 image for different poses and expressions (4674-2D, 4674-3D Images. During building the database, not only the single variations of poses, but also expressions and illuminations are considered.

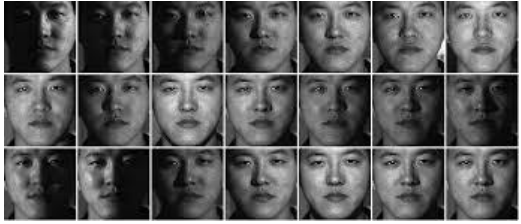

Fig. 6 CASIA 2D face Database

### ORL Database
It contains a set of faces taken between April 1992 and April 1994 at the Olivetti Research Laboratory in Cambridge, K. There are 10 different images of 40 distinct subjects. For some of the subjects, the images were taken at different situations, varying lighting slightly, facial expressions open/closed eyes, smiling/non-smiling) and facial details (glasses/no-glasses) [15].

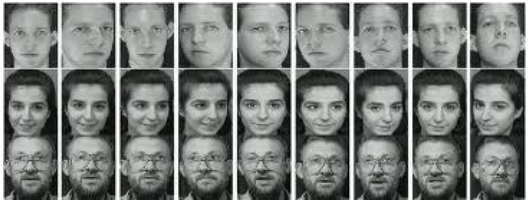

Fig. 7 ORL 2D face Database

### Cropped YaleB Database
The database contains 2535 Portable Graphics Media (pgm) images of 39 subjects each seen under different viewing conditions with different poses and different illumination conditions. For every subject in a particular pose, an image with ambient (background) illumination was also captured.

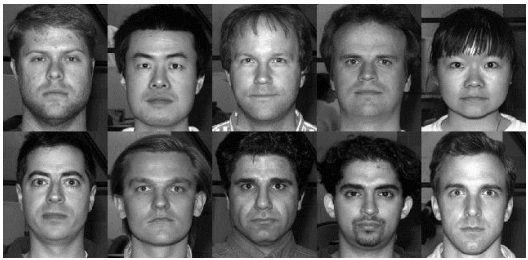

Fig. 8 Extended YaleB 2D face Database

The databases have been organized as follows: a training set of 4 images, a verification set of 3 images and finally 3 images for testing (for each subject).

The algorithm of the proposed method can be summarized as:
- Preprocessing the face image by finding the locations of eyes centers then crop and rotate the face image.
- Adaptive Single Scale Retinex has been used for illumination normalization.
- Select orthogonal and non-correlated subset Gabor filters to extract the face features using PCA.
- Feature extraction of the face images using the orthogonal Gabor filters.
- Feature vector compression using LDA.
- Classification of the generated compressed features and testing the performance.

The proposed system has achieved high recognition rates compared to current systems as shown in table-1. Since, the storage space and computations are as important as the accuracy of verification and/or identification. The proposed method has achieved these recognition rates using only 25 orthogonal Gabor filters instead of 40 Gabor filters (62.5 % of the original filters) which reduce the time of calculation and storage needed to store the generated face database.

| Data base | Proposed Sys. Rec. Rate | Current Sys. Rec. Rates |
|---|---|---|
| CASIA | 99.17% | 99.50 %[24] |
| ORL | 98.33% | 97 %[25] |
| Cropped YaleB | 99.33% | 99.16%[26] |

Table 1: Recognition Rates of the orthogonal filters

## 6. Conclusion

In this paper we have presented a robust, fast highly accurate 2D face recognition system based on feature extraction using best selected orthogonal Gabor filters which decreased the feature vector length and reduced the processing time. The system has been tested using three 2d Data Bases CASIA Database, ORL Database and Cropped YaleB Database with an enhancement to the input 2D face image using multi stage preprocessing technique to normalize and enhance the input image. The overall recognition results achieved demonstrate the significant effective performance of the proposed system.

**Mazen M. Selim** received the BSc in Electrical Engineering in 1982, the MSc in 1987 and PhD in 1993 from Zagazig University (Benha Branch) in electrical and communication engineering. He is now an Associate Professor at the faculty of computers and informatics, Benha University. His areas of interest are image processing, biometrics, sign language, content based image retrieval (CBIR), face recognition and watermarking.

**Hala H. Zayed** received the BSc in Electrical Engineering (with honor degree) in 1985, the MSc in 1989 and PhD in 1995 from Zagazig University (Benha Branch) in electrical and communication engineering. She is now a professor at faculty of computers and informatics, Benha University. Her areas of research are pattern recognition, content based image retrieval, biometrics and image processing.

**Samir F. Hafez** received the BSc in Electrical Engineering (with honor degree) in 1996 from Zagazig University (Benha Branch) and the MSc in 2006 from Benha University in Electrical and Communication Engineering.